\documentclass[letterpaper, 10 pt, conference]{ieeeconf}  

\IEEEoverridecommandlockouts                              
\usepackage{graphicx}   
\usepackage{adjustbox}  
\usepackage{graphicx}   
\usepackage{multirow}   
\usepackage{booktabs}   
\usepackage{amsmath}    
\usepackage{amssymb} 
\usepackage{bm}
\usepackage{pifont}
\usepackage{xcolor}

\usepackage{float}
\usepackage{xcolor}
\usepackage[
    colorlinks=true,
    urlcolor=magenta     
]{hyperref}

\title{\LARGE \bf
THAT: Token-wise High-frequency Augmentation Transformer for Hyperspectral Pansharpening
}

\author{Hongkun Jin $^{1}$\textsuperscript{*}, Hongcheng Jiang, Graduate Student Member, IEEE $^{2}$\textsuperscript{*}, Zejun Zhang, Graduate Student \\Member, IEEE $^{3}$\textsuperscript{*}, Yuan Zhang $^{4}$, Jia Fu, Graduate Student Member, IEEE $^{5}$, Tingfeng Li $^{6}$, Kai Luo $^{7}$\textsuperscript{\dag}
\thanks{\textsuperscript{*}Equal contribution}
\thanks{\textsuperscript{\dag}Corresponding author}
\thanks{$^{1}$Hongkun Jin is with of JPMorgan Chase, 8181 Communications Pkwy Building F, Plano, TX 75024, USA. He was with the Electrical Computer Engineering Department, University of Missouri-Kansas City, Kansas City, MO 64111, USA. 
        {\tt\small max.jin@chase.com}}%
\thanks{$^{2}$Hongcheng Jiang is with the Electrical Computer Engineering Department, University of Missouri-Kansas City, Kansas City, MO 64111, USA. 
        {\tt\small hjq44@mail.umkc.edu}}%
 \thanks{$^{3}$Zejun Zhang is with  Ming Hsieh Department of Electrical and Computer Engineering, University of Southern California, Los Angeles, CA 90007, USA. 
        {\tt\small zejunzha@usc.edu}}%
 \thanks{$^{4}$Yuan Zhang is with Robinson Research Institute, University of Adelaide, Adelaide, SA 5000, AU. 
        {\tt\small yuan.zhang01@adelaide.edu.au}}%
 \thanks{$^{5}$Jia Fu is with KTH Royal Institute of Technology, Stockholm, 114 28, SE. 
        {\tt\small  jiafu@kth.se}}%
 \thanks{$^{6}$Tingfeng Li is with NEC Laboratories America, Princeton, NJ 08540, USA. 
        {\tt\small tli@nec-labs.com}}
 \thanks{$^{7}$Kai Luo was with University of Virginia, Charlottesville, VA 22904, USA. 
        {\tt\small  kl3pq@virgina.com}}%
}

\begin{document}

\maketitle
\thispagestyle{empty}
\pagestyle{empty}

\begin{abstract}

Transformer-based methods have demonstrated strong potential in hyperspectral pansharpening by modeling long-range dependencies. However, their effectiveness is often limited by redundant token representations and a lack of multi-scale feature modeling. Hyperspectral images exhibit intrinsic spectral priors (e.g., abundance sparsity) and spatial priors (e.g., non-local similarity), which are critical for accurate reconstruction. From a spectral--spatial perspective, Vision Transformers (ViTs) face two major limitations: they struggle to preserve high-frequency components---such as material edges and texture transitions---and suffer from attention dispersion across redundant tokens. These issues stem from the global self-attention mechanism, which tends to dilute high-frequency signals and overlook localized details. To address these challenges, we propose the \textit{Token-wise High-frequency Augmentation Transformer (THAT)}, a novel framework designed to enhance hyperspectral pansharpening through improved high-frequency feature representation and token selection. Specifically, THAT introduces: (1) \textit{Pivotal Token Selective Attention (PTSA)} to prioritize informative tokens and suppress redundancy; (2) a \textit{Multi-level Variance-aware Feed-forward Network (MVFN)} to enhance high-frequency detail learning. Experiments on standard benchmarks show that THAT achieves state-of-the-art performance with improved reconstruction quality and efficiency. The source code is available at \url{https://github.com/kailuo93/THAT}.

\end{abstract}


\section{Introduction}

Hyperspectral imaging captures rich spectral information by acquiring spatially distributed spectral profiles, where each profile represents the reflectance or radiance of a pixel across specific wavelengths. This capability facilitates material identification and supports various remote sensing applications, including classification~\cite{9815317}, spectral unmixing~\cite{duan2023undat}, and segmentation~\cite{hang2022multiscale}. However, achieving advanced visual understanding in hyperspectral images (HSIs), such as semantic object recognition, necessitates high spatial resolution comparable to color imagery. Hyperspectral imaging systems inherently face a spectral-spatial trade-off~\cite{jia2021tradeoffs}, where high spectral resolution is obtained at the cost of spatial resolution due to limited light throughput in narrow-band optical filtering~\cite{armin2015narrowband} and cost-driven constraints in sensor miniaturization~\cite{stuart2020low}. A practical and cost-effective approach to overcoming these limitations is hyperspectral pansharpening, which integrates low-resolution HSIs (LR-HSIs) with high-resolution panchromatic images (HR-PCIs) to reconstruct high-resolution HSIs (HR-HSIs) with improved spatial and spectral fidelity~\cite{thomas2008synthesis, loncan2015hyperspectral}.

\begin{figure}[tbp!]
    \centering
    \resizebox{\columnwidth}{!}{
        \includegraphics[height=4.5cm]{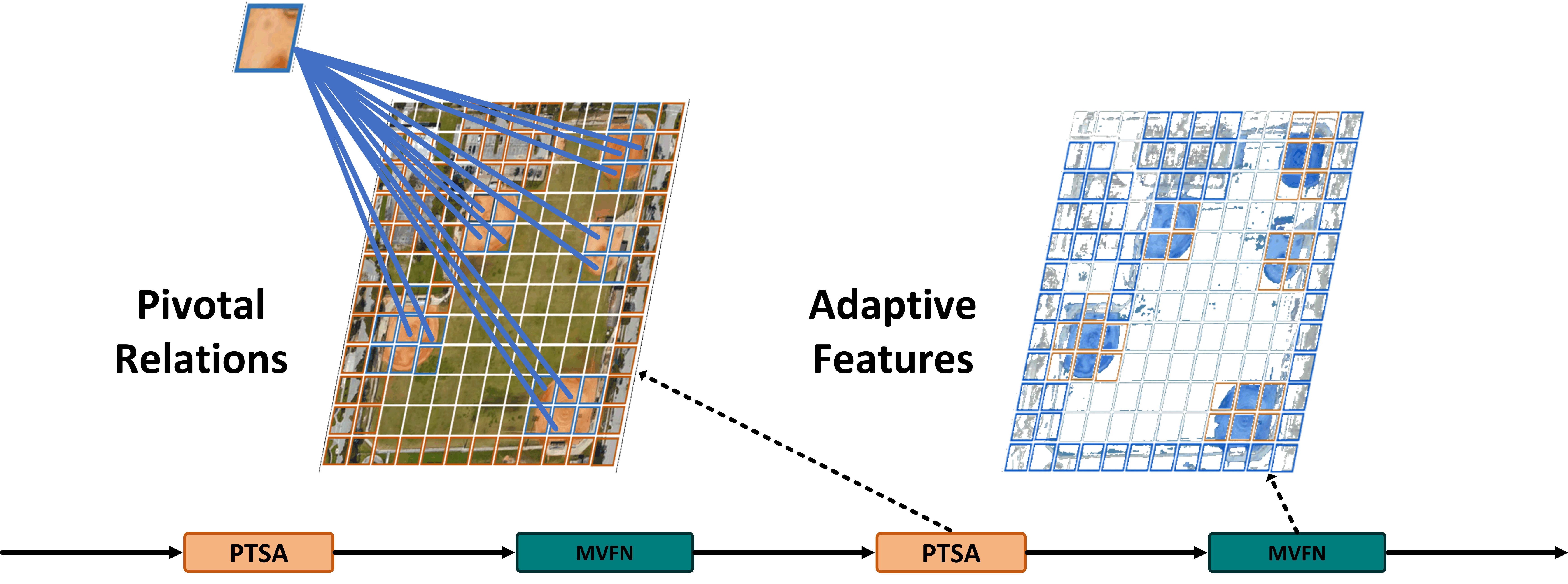}
    }
    \caption{Feature visualization of the proposed Token-wise High-frequency Augmentation Transformer (THAT), which integrates two key modules: (1) Pivotal Token Selective Attention (PTSA), designed to identify and emphasize informative tokens while suppressing less relevant ones to improve attention efficiency; and (2) Multi-level Variance-aware Feed-forward Network (MVFN), which captures hierarchical spectral–spatial dependencies to explicitly enhance high-frequency detail learning.}

    \label{fig:com}
\end{figure}

Existing hyperspectral pansharpening techniques can be broadly categorized into two groups: statistical modeling-based approaches~\cite{aly2014regularized, dong2019improved} and machine learning-based approaches~\cite{guarino2024hybrid}. The former typically adopts unsupervised estimation strategies by formulating the inverse imaging problem as an optimization task. These methods are further classified into four main categories~\cite{loncan2015hyperspectral}: component substitution, multi-resolution analysis, Bayesian inference, and matrix factorization. While these approaches are generally computationally efficient, they often introduce spectral distortions during HR-HSI reconstruction~\cite{wang2021hyperspectral}. In contrast, machine learning-based methods—particularly deep learning approaches—have shown promising results in hyperspectral pansharpening, owing to their powerful feature representation capabilities. Convolutional neural networks (CNNs) have been widely adopted to model the nonlinear relationship between LR-HSIs and HR-HSIs in an end-to-end manner~\cite{he2023dynamic}. More recently, transformer-based architectures have gained attention due to their multi-head self-attention mechanisms, which enable better modeling of long-range dependencies and global context compared to CNNs. However, existing transformer-based methods typically rely on dense self-attention for feature aggregation, where all tokens are considered for similarity computation, without accounting for the unique spectral–spatial characteristics of hyperspectral data.

Transformer-based methods have shown strong potential in hyperspectral pansharpening by effectively modeling long-range dependencies \cite{bandara2022hypertransformer}. However, their performance is often limited by redundant token representations and the lack of multi-scale feature modeling. Hyperspectral images exhibit intrinsic spectral priors (e.g., abundance sparsity) and spatial priors (e.g., non-local similarity), both essential for accurate spectral--spatial reconstruction. From this perspective, Vision Transformers (ViTs) face two key limitations: difficulty in preserving high-frequency components—such as material edges and texture transitions—and dispersion of attention across redundant tokens. These challenges reflect two core issues: spectral--spatial inconsistency,and spectral redundancy. The former arises from single-scale global modeling, which tends to blur fine details and introduces spatial artifacts. The latter stems from strong spectral correlations, leading to redundant token representations that dilute attention across both informative and uninformative regions. These limitations ultimately hinder the ability of conventional transformer-based methods to preserve spectral fidelity and spatial detail in hyperspectral pansharpening.

In the last years, Transformer-based methods have been extensively explored to address the challenges of redundant token representations and single-scale modeling in super-resolution (SR) tasks. Xiao et al. \cite{xiao2024ttst} proposed the Top-k Token Selective Transformer (TTST) for remote sensing image SR, which effectively refines the attention mechanism by selecting the most relevant tokens. However, this approach is computationally expensive as it requires multiple iterations to determine effective tokens, and the selection ratio significantly influences the computed Top-k tokens, affecting stability. Zhou et al. \cite{zhou2024adapt} introduced a ReLU-based Sparse Self-Attention (SSA) from Natural Language Processing (NLP) to filter out noisy interactions among irrelevant tokens. While this method prevents information loss due to small entropy in SSA, it does not account for spatial relationships between neighboring tokens, limiting its ability to model local dependencies. Additionally, Jiang et al. \cite{jiang2024flexible} developed a Flexible Window-based Self-Attention Transformer (FW-SAT) tailored for thermal image super-resolution. Despite its ability to handle varying spatial resolutions dynamically, FW-SAT incurs high memory costs due to the computational overhead of flexible pointer calculations.

\begin{figure}[htbp]
    \centering
    \resizebox{\columnwidth}{!}{
        \includegraphics{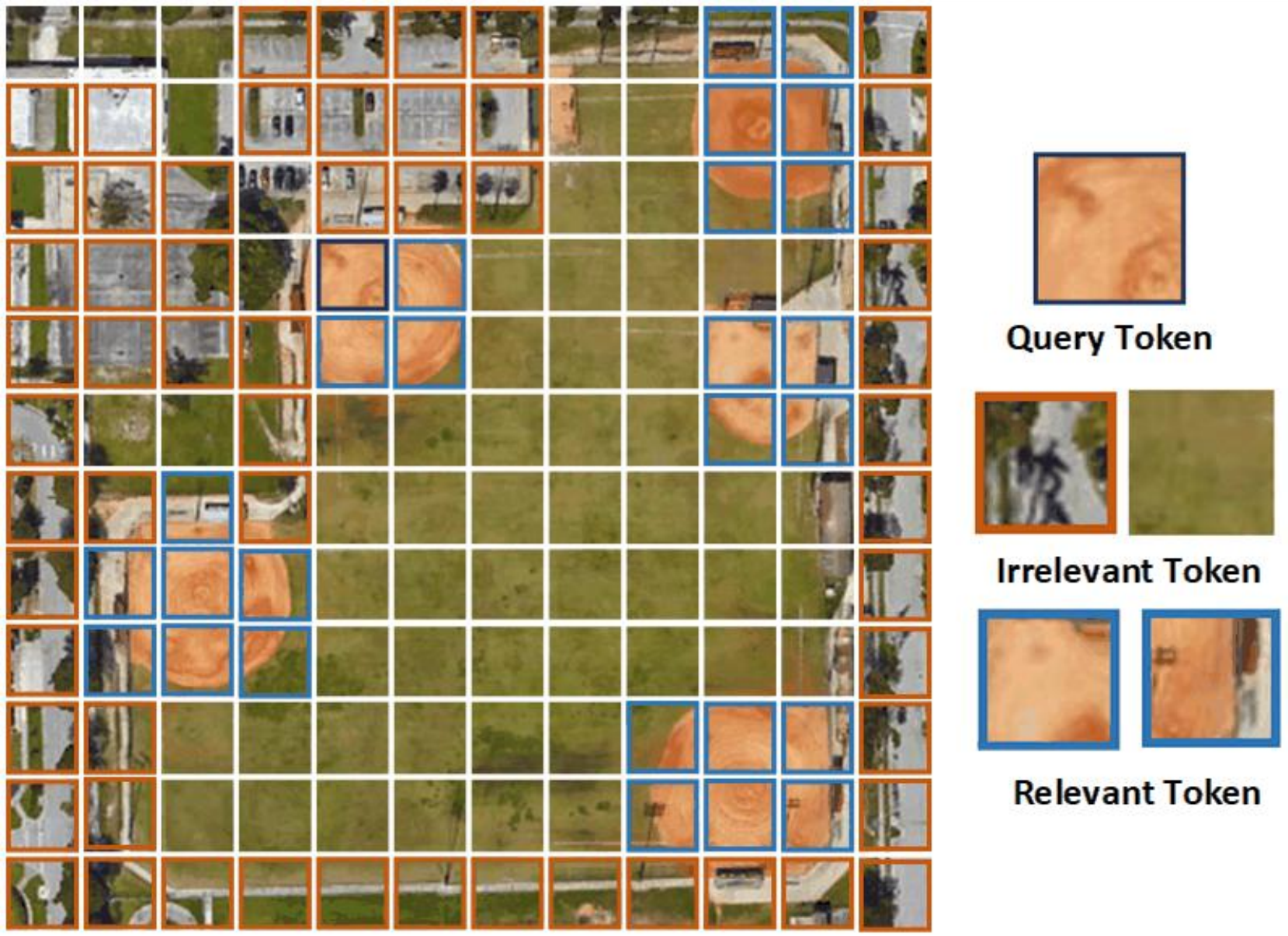}
    }
    \caption{
        Illustration of token selection in the proposed THAT for hyperspectral pansharpening. The figure highlights the limitations of traditional transformer-based approaches, which suffer from redundant token representations and inefficient single-scale modeling. THAT addresses these issues through PTSA, dynamically refining self-attention by prioritizing informative tokens (blue) while filtering out redundant ones (orange). This enables more effective spectral-spatial correlation modeling. Additionally, the MVFN enhances hierarchical feature aggregation. The right side of the figure shows query, relevant, and irrelevant tokens, illustrating THAT's token selection mechanism.
    }
    \label{fig:example}
\end{figure}

To overcome these challenges, we propose Token-wise High-frequency Augmentation Transformer (THAT). The first component, Pivotal Token Selective Attention (PTSA), dynamically prioritizes informative tokens while filtering redundant ones. PTSA leverages k-means clustering, offering several advantages, including efficiency and scalability, simple and easy implementation, and flexibility in application. Unlike existing Transformer-based methods that suffer from computational redundancy and single-scale limitations, PTSA ensures that only the most relevant spectral-spatial features contribute to the reconstruction process, significantly improving efficiency while preserving structural integrity.

We further introduce the Multi-level Variance-aware Feed-forward Network (MVFN) to explicitly enhance high-frequency detail learning by capturing hierarchical spectral--spatial dependencies. Unlike conventional feed-forward networks, MVFN models inter-token variance across multiple levels to adaptively respond to varying spectral complexity. This variance-aware mechanism significantly improves both spectral fidelity and spatial detail reconstruction in the fused hyperspectral output. By integrating PTSA and MVFN with the feature visualization shown in Fig.~\ref{fig:com}, our method achieves state-of-the-art performance in hyperspectral pansharpening, delivering a compelling balance of accuracy, efficiency, and scalability.

\begin{figure*}[htbp]
    \centering
    \includegraphics[width=\textwidth]{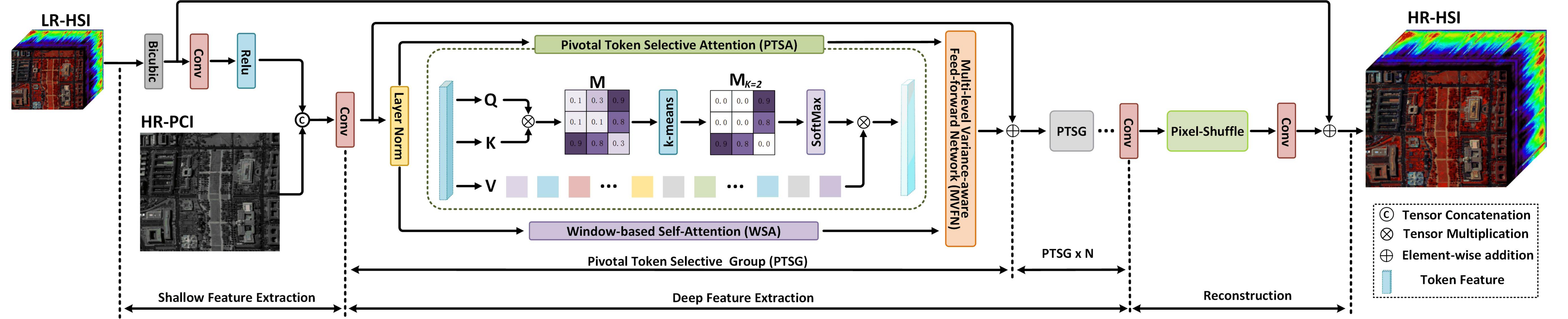}
    \caption{
    Overview of the proposed THAT architecture for hyperspectral pansharpening.
    }
    \label{fig:Network}
\end{figure*}

The main contributions of this paper are summarized as follows:

\begin{itemize}
\item We propose the \textit{Token-wise High-frequency Augmentation Transformer (THAT)}, a novel framework for hyperspectral pansharpening that addresses token redundancy and enhances high-frequency feature representation. THAT effectively integrates spectral–spatial dependencies, leading to improved reconstruction with superior spectral fidelity and spatial sharpness.

\item We introduce a \textit{Pivotal Token Selective Attention (PTSA)} module that dynamically identifies and emphasizes informative tokens while suppressing redundant ones. This selective mechanism improves the efficiency of self-attention and boosts the discriminative power of token representations for hyperspectral fusion.

\item We design a \textit{Multi-level Variance-aware Feed-forward Network (MVFN)} to explicitly enhance high-frequency detail learning. By modeling spectral–spatial variance across multiple levels, MVFN significantly improves both spectral preservation and spatial detail reconstruction in the fused hyperspectral output.

\end{itemize}

\section{Related Work}

\subsection{Statistical Estimation Methods}
Traditional hyperspectral pansharpening techniques primarily rely on statistical modeling-based approaches, which reconstruct HR-HSIs from low-resolution inputs via mathematical formulations. Component Substitution (CS) methods replace spatial details in LR-HSIs with HR-PCI features~\cite{thomas2008synthesis}, but often introduce spectral distortions~\cite{rui2024unsupervised}. Multi-Resolution Analysis (MRA) methods enhance resolution by injecting multi-scale spatial details from HR-PCIs~\cite{loncan2015hyperspectral}, though they suffer from aliasing effects~\cite{vivone2018full}. Bayesian estimation formulates pansharpening as an inverse problem, modeling spectral priors for robust reconstruction~\cite{simoes2014convex}, while variational methods impose regularization constraints to balance fidelity and prior information~\cite{ballester2006variational}. Despite their efficiency, these methods struggle with handcrafted priors, limited spectral-spatial modeling, leading to the rise of data-driven deep learning approaches.

\subsection{Machine Learning Methods}
Machine learning-based hyperspectral pansharpening approaches can be broadly categorized into supervised and unsupervised methods, with deep learning (DL) emerging as the dominant paradigm since 2015. Supervised DL methods leverage large-scale training data to learn nonlinear mappings between LR-HSIs and HR-HSIs. Xu et al. \cite{xu2021deep} proposed Deep Gradient Projection Networks (DGPNet), integrating iterative gradient projection steps to refine the fused output while preserving spectral fidelity. Qu et al. \cite{qu2021dual} introduced the Dual-Branch Detail Extraction Network (DBDEN), which captures both spectral and spatial information to enhance fine-detail preservation. Guan and Lam \cite{guan2021multistage} developed the Multistage Dual-Attention Guided Fusion Network (MDAGFN), which utilizes spatial and spectral attention mechanisms to achieve superior fusion quality. However, DL methods face challenges such as computational inefficiency, spectral redundancy, and limited interpretability. Recently, transformer-based models have been explored for their long-range dependency modeling, but dense self-attention fails to address spectral redundancy and spatial inconsistencies. Efficient transformer-based approaches are needed to overcome these limitations.

\section{Proposed Method} 

\subsection{Overall Pipeline}
As illustrated in Fig.~\ref{fig:Network}, the proposed Token-wise High-frequency Augmentation Transformer (THAT) follows a three-stage architecture comprising shallow feature extraction, deep feature extraction, and feature reconstruction, a widely adopted structure in prior works~\cite{chen2023activating,jiang2024flexible}. Given a LR-HSI \( \bm{Y} \in \mathbb{R}^{h \times w \times S} \) and a HR-PCI \( \bm{X} \in \mathbb{R}^{H \times W} \), where \( S \) denotes the number of spectral bands, bicubic interpolation is first applied to \( \bm{Y} \), followed by a convolutional layer with ReLU activation to extract shallow features, which are then concatenated with HR-PCI for spatial guidance. The deep feature extraction stage leverages the Pivotal Token Selective Group (PTSG), which consists of PTSA for dynamically prioritizing informative tokens while filtering redundancy, Window-based Self-Attention (WSA) for capturing local spectral-spatial interactions, and the MVFN to model hierarchical spectral-spatial dependencies for feature enhancement. The PTSG is stacked \( N \) times to progressively refine feature representations. Finally, the feature reconstruction stage employs a convolutional layer followed by a pixel-shuffle operation to upsample the fused features, reconstructing the target HR-HSI \( \bm{F}_t \in \mathbb{R}^{H \times W \times S} \), ensuring robust spectral-spatial consistency.

\begin{figure}[tbp!]
    \centering
    \includegraphics[width=\columnwidth]{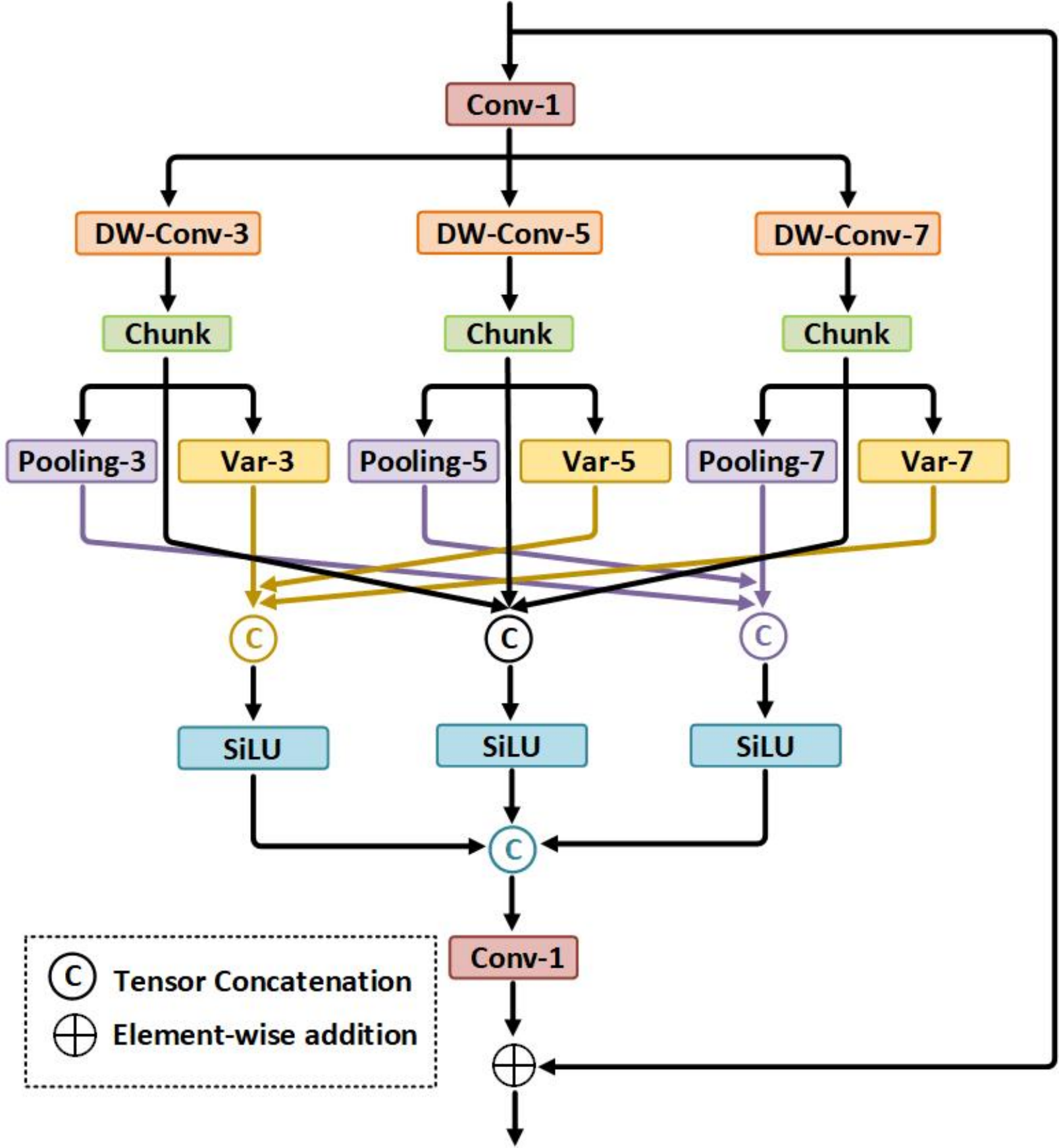}
\caption{Structure of the MVFN, designed to capture hierarchical spectral–spatial dependencies and enhance high-frequency feature representation. MVFN leverages multi-scale depthwise convolutions, variance modeling, and adaptive feature aggregation to refine both spectral and spatial details.}
    \label{fig:mvfn}
\end{figure}

\subsubsection{Pivotal Token Selective Attention (PTSA)}
PTSA refines self-attention by dynamically selecting and prioritizing informative tokens while suppressing redundant ones. As shown in Fig.~\ref{fig:Network}, given query \( Q \), key \( K \), and value \( V \) representations, PTSA first computes the raw attention matrix:

\begin{equation}
    M = (QK^T) \cdot \tau,
\end{equation}

where \( \tau \) is a learnable temperature parameter that scales the dot product operation, improving numerical stability. Instead of applying SoftMax directly to all token pairs, PTSA introduces a k-means clustering step to partition tokens into pivotal and non-pivotal groups. The k-means algorithm clusters tokens based on similarity scores in \( M \), enabling the model to focus on essential spectral-spatial interactions. Specifically, the k-means algorithm clusters tokens into two groups based on their similarity scores in \( M \). The cluster with larger average similarity values is considered pivotal, as it reflects stronger spectral-spatial interactions. A binary mask is then applied: tokens in the high-value (pivotal) cluster are assigned a mask value of 1, while those in the low-value cluster are set to 0. This mask is used to filter the attention matrix, yielding a refined attention map \( M' \) that focuses on the most informative interactions and suppresses less relevant ones. The refined attention matrix \( M' \) is then computed by filtering out non-pivotal tokens:

\begin{equation}
    M' = \text{k-means}(M).
\end{equation}

After clustering, PTSA applies a SoftMax operation only to the pivotal tokens:

\begin{equation}
    A = \text{SoftMax}(M').
\end{equation}

The final attention output is then computed by applying the attention weights to the value matrix:

\begin{equation}
    O = A V.
\end{equation}

To further regulate token selection, PTSA normalizes the query and key features before computing attention:

\begin{equation}
    Q' = \frac{Q}{\|Q\|}, \quad K' = \frac{K}{\|K\|}.
\end{equation}

This ensures stable similarity computation and prevents large-scale variations in feature magnitudes. Finally, the output is projected back to the original feature space using a convolutional layer:

\begin{equation}
    O' = \text{Conv}(O).
\end{equation}

By integrating k-means clustering for token selection, feature normalization, and selective self-attention, PTSA enhances hyperspectral image fusion by prioritizing relevant tokens while filtering out irrelevant ones, effectively capturing essential spectral-spatial dependencies and improving computational efficiency.

\subsubsection{Multi-Level Variance-aware Feed-forward Network (MVFN)}
MVFN enhances hyperspectral feature representations by focusing on high-frequency spectral-spatial details. As illustrated in Fig.~\ref{fig:mvfn}, MVFN incorporates multi-scale depthwise convolutions (DW-Conv-3, DW-Conv-5, DW-Conv-7) to extract spatial features across varying receptive fields, which is crucial for detecting high-frequency textures and edges. Following each convolutional path, variance modeling modules (Var-3, Var-5, Var-7) estimate local statistical variances, allowing the network to emphasize subtle, high-frequency variations critical for preserving spectral fidelity and spatial sharpness. To further suppress redundant low-frequency components and refine salient details, pooling operations are employed within each branch. The outputs from all branches are then aggregated using concatenation and element-wise addition, followed by a SiLU activation and a final convolution layer to unify the enriched features. This hierarchical and frequency-aware design enables MVFN to selectively amplify high-frequency information, thereby improving the reconstruction quality of fine-grained spectral structures and enhancing spatial clarity.

\begin{table}[t]
\caption{Quantitative results for HSI pansharpening ($\times2$). \textbf{Best} and \underline{second-best} values are highlighted.}
\centering
\resizebox{\columnwidth}{!}{
\begin{tabular}{c|c|c|c|c|c|c}
\hline
\textbf{Dataset} & \textbf{Method}  & \textbf{PSNR} $\uparrow$ & \textbf{SSIM} $\uparrow$ & \textbf{SAM} $\downarrow$ & \textbf{ERGAS} $\downarrow$ & \textbf{SCC} $\uparrow$ \\ 
\hline
              & DBDENet \cite{qu2021dual}   & 19.51  & 0.3862  & 11.9602  & 11.8282  & 0.5874  \\
              & DDLPS  \cite{li2019ddlps}  & 22.45  & 0.5934  & 6.0663  & 18.1751  & 0.8402  \\   
              & DHP-DARN \cite{zheng2020hyperspectral}  & 27.64  & 0.8527  & 3.3296  & 2.8252  & 0.9189 \\
              & DIP-HyperKite \cite{gedara2021hyperspectral}  & 28.69  & 0.8510  & 3.5377  & 2.1638  & 0.8957  \\
              & DMLD-Net \cite{zhang2023learning}  & 21.63  & 0.7407  & 7.5993  & 5.2205  & 0.8765  \\
\textbf{Botswana} & GPPNN \cite{xu2021deep} & 22.83 & 0.7188 & 11.2243 & 4.3593 & 0.8410 \\
              & GS \cite{laben2000process}  & 10.31 & 0.4967 & 15.0426 & 64.1328 & 0.8502 \\
              & GSA \cite{aiazzi2007improving}  & 26.44 & 0.7221 & 4.0392 & 8.8435 & 0.9030 \\
              & Indusion \cite{khan2008indusion}  & 15.27 & 0.7097 & \underline{3.2173} & 9.8120 & 0.9036 \\
              & PLRDiff \cite{rui2024unsupervised}   & 15.27  & 0.3246  & 17.0449  & 14.2717  & 0.4845  \\ 
              & PSDip \cite{rui2024variational}  & \textbf{29.20} & 0.8755 & 4.7042 & \underline{2.0956} & 0.8944 \\
              & TTST \cite{xiao2024ttst}  & 28.07  & \underline{0.8877}   & 3.2453   & 2.4024   & \underline{0.9369}  \\ 
              & \textbf{Ours}  & \underline{29.18} & \textbf{0.9084} & \textbf{2.6657}  & \textbf{2.0754} & \textbf{0.9493} \\ 
\hline
              & DBDENet \cite{qu2021dual}   & 27.95  & 0.8326  & 10.0722  & 4.4037  & 0.9046  \\
              & DDLPS  \cite{li2019ddlps}  & 29.41  & 0.8474  & 11.7790  & 4.0209  & 0.9049  \\  
              & DHP-DARN \cite{zheng2020hyperspectral}  & 31.60  & 0.9014  & 7.6660  & 2.8416  & 0.9327  \\
              & DIP-HyperKite \cite{gedara2021hyperspectral}  & 34.33  & \underline{0.9536}  & \underline{5.3473}  & 2.1306  & 0.9734  \\
              & DMLD-Net \cite{zhang2023learning} & 29.41  & 0.8779  & 8.0543  & 3.7124  & 0.9456  \\
\textbf{PaviaC} & GPPNN \cite{xu2021deep}  & 30.88  & 0.9009  & 8.1573  & 3.2473  & 0.9578  \\
              & GS \cite{laben2000process} & 31.75  & 0.8974  & 7.9654  & 3.0498  & 0.9291  \\
              & GSA \cite{aiazzi2007improving} & 30.18  & 0.8857  & 7.7252  & 3.4145  & 0.9205  \\
              & Indusion \cite{khan2008indusion} & 32.54  & 0.9356  & 6.7197  & 2.7776  & 0.9481  \\
              & PLRDiff \cite{rui2024unsupervised}  & 33.45  & 0.9362  & 7.8851  & 2.5256  & 0.9690  \\ 
              & PSDip \cite{rui2024variational}  & 27.75  & 0.8867  & 8.7329  & 4.3015  & 0.8115  \\
              & TTST \cite{xiao2024ttst}  & \underline{34.98}  & 0.9529  & 5.6894  & \underline{1.9811}  & \underline{0.9774}  \\
              & \textbf{Ours}  & \textbf{35.29}  & \textbf{0.9574}  & \textbf{5.2818}  & \textbf{1.8838}  & \textbf{0.9792}  \\
\hline
              & DBDENet \cite{qu2021dual}   & 29.79  & 0.8853  & 6.4053  & 2.4789  & 0.9229  \\
              & DDLPS  \cite{li2019ddlps}  & 30.87  & 0.8931  & 6.5185  & 2.2404  & 0.9148  \\ 
              & DHP-DARN \cite{zheng2020hyperspectral}  & 30.87  & 0.8931  & 6.5185  & 2.2404  & 0.9148  \\
              & DIP-HyperKite \cite{gedara2021hyperspectral}  & 35.55  & 0.9495  & 3.4424  & 1.2701  & 0.9736  \\
              & DMLD-Net \cite{zhang2023learning}  & 30.81  & 0.9003  & 5.7911  & 2.2189  & 0.9487  \\
\textbf{PaviaU} & GPPNN \cite{xu2021deep} & 33.46  & 0.9362  & 4.8439  & 1.6055  & 0.9641  \\
              & GS \cite{laben2000process}  & 33.43  & 0.9186  & 5.0129  & 1.6940  & 0.9376  \\
              & GSA \cite{aiazzi2007improving}  & 32.17  & 0.9052  & 4.8603  & 1.8747  & 0.9324  \\
              & Indusion \cite{khan2008indusion}  & 34.09  & 0.9313  & 4.5238  & 1.5609  & 0.9537  \\
              & PLRDiff \cite{rui2024unsupervised}   & 35.33  & 0.9420  & 4.6869  & 1.4236  & 0.9740  \\             
              & PSDip \cite{rui2024variational}  & 31.16  & 0.8893  & 6.0024  & 1.9748  & 0.9076  \\
              & TTST \cite{xiao2024ttst}  & \underline{37.35}  & \underline{0.9618}  & \underline{3.2400}  & \underline{1.0775}  & \textbf{0.9818}  \\
              & \textbf{Ours} & \textbf{37.82}  & \textbf{0.9632}  & \textbf{3.0172}  & \textbf{1.0039}  & \underline{0.9816}  \\
\hline
\end{tabular}
}
\label{tab:2_20}
\end{table}

\begin{table}[ht]
\caption{Quantitative results for HSI pansharpening ($\times4$). \textbf{Best} and \underline{second-best} values are highlighted.}
\centering
\resizebox{\columnwidth}{!}{
\begin{tabular}{c|c|c|c|c|c|c}
\hline
\textbf{Dataset} & \textbf{Method}  & \textbf{PSNR} $\uparrow$ & \textbf{SSIM} $\uparrow$ & \textbf{SAM} $\downarrow$ & \textbf{ERGAS} $\downarrow$ & \textbf{SCC} $\uparrow$ \\ 
\hline
      & DBDENet \cite{qu2021dual}   & 24.78  & 0.8108  & 5.2043  & 3.7034  & 0.8814  \\
      & DDLPS  \cite{li2019ddlps}  & 22.91  & 0.5997  & 6.2523  & 17.6875  & 0.7851  \\     
      & DHP-DARN \cite{zheng2020hyperspectral}  & \textbf{29.64}  & 0.8482  & 4.1249  & 2.5237  & 0.8911  \\
      & DIP-HyperKite \cite{gedara2021hyperspectral}  & 29.32  & 0.8592  & 4.2999  & \underline{2.1662}  & 0.8929  \\
      & DMLD-Net \cite{zhang2023learning}  & 25.35  & 0.8069  & 5.2349  & 3.5039  & 0.8797  \\
\textbf{Botswana} & GPPNN \cite{xu2021deep} & 26.59  & 0.8419  & 6.8525  & 2.9596  & 0.8793  \\
      & GS \cite{laben2000process}  & 10.25  & 0.4824  & 15.3088  & 65.2786  & 0.7935  \\
      & GSA \cite{aiazzi2007improving}  & 24.63  & 0.6660  & 5.2871  & 10.3319  & 0.8301  \\
      & Indusion \cite{khan2008indusion}  & 15.29  & 0.7038  & 4.4775  & 9.7700  & 0.8714  \\
      & PLRDiff \cite{rui2024unsupervised}   & 19.71  & 0.5255  & 13.2378  & 8.1385  & 0.5819  \\          
      & PSDip \cite{rui2024variational}  & 29.10  & \textbf{0.8756}  & 4.7030  & \textbf{2.0956}  & \underline{0.8945}  \\
      & TTST \cite{xiao2024ttst}  & 29.00  & 0.8543  & \underline{4.0126}  & 2.3589  & 0.8885  \\
      & \textbf{Ours}  & \underline{29.34}  & \underline{0.8728}  & \textbf{3.8377}  & 2.3373  & \textbf{0.8979}  \\
\hline
      & DBDENet \cite{qu2021dual}   & 28.59  & 0.8347  & 9.5381  & 4.0255  & 0.8948  \\
      & DDLPS  \cite{li2019ddlps}  & 29.24  & 0.8375  & 10.1167  & 3.7741  & 0.8716  \\            
      & DHP-DARN \cite{zheng2020hyperspectral}  & 31.06  & 0.8940  & 8.1013  & 3.0066  & 0.9246  \\
      & DIP-HyperKite \cite{gedara2021hyperspectral}  & 29.69  & 0.8667  & \underline{8.0389}  & 3.5091  & 0.9178  \\
      & DMLD-Net \cite{zhang2023learning}  & 28.32  & 0.8420  & 9.3755  & 4.0618  & 0.9008  \\
\textbf{PaviaC} & GPPNN \cite{xu2021deep}  & 28.77  & 0.8469  & 10.7878  & 3.9314  & 0.9072  \\
      & GS \cite{laben2000process} & 29.93  & 0.8343  & 10.5851  & 3.7002  & 0.8769  \\
      & GSA \cite{aiazzi2007improving} & 27.39  & 0.8015  & 9.5616  & 4.5898  & 0.8510  \\
      & Indusion \cite{khan2008indusion} & 30.97  & 0.8974  & 8.6969  & 3.2146  & 0.9154  \\
      & PLRDiff \cite{rui2024unsupervised}  & 31.28  & 0.8881  & 9.7650  & 3.1999  & 0.9178  \\
      & PSDip \cite{rui2024variational}  & 24.45  & 0.8188  & 10.5988  & 6.2748  & 0.6533  \\
      & TTST \cite{xiao2024ttst}  & \underline{31.97}  & \underline{0.9044}  & 8.1011  & \underline{2.7302}  & \underline{0.9357}  \\
      & \textbf{Ours}  & \textbf{32.43}  & \textbf{0.9157}  & \textbf{7.4978}  & \textbf{2.5841}  & \textbf{0.9420}  \\
\hline
      & DBDENet \cite{qu2021dual}    & 25.99  & 0.8396  & 8.5022  & 3.9397  & 0.8571  \\
      & DDLPS  \cite{li2019ddlps} & 30.29  & 0.8642  & 6.4812  & 2.2603  & 0.8846  \\          
      & DHP-DARN \cite{zheng2020hyperspectral}  & 31.45  & 0.8926  & 5.5492  & 1.9958  & 0.9169  \\
      & DIP-HyperKite \cite{gedara2021hyperspectral}  & 30.27  & 0.8769  & 5.8648  & 2.2594  & 0.9094  \\
      & DMLD-Net \cite{zhang2023learning} & 28.11  & 0.8575  & 6.9105  & 2.9301  & 0.8839  \\
\textbf{PaviaU} & GPPNN \cite{xu2021deep} & 28.52  & 0.8675  & 7.0388  & 2.8360  & 0.8990  \\
      & GS \cite{laben2000process}  & 31.25  & 0.8695  & 6.7026  & 2.1306  & 0.8899  \\
      & GSA \cite{aiazzi2007improving}  & 29.01  & 0.8368  & 6.2704  & 2.6495  & 0.8720  \\
      & Indusion \cite{khan2008indusion}  & 31.69 & 0.8869  & 6.1201  & 2.0051  & 0.9131  \\
      & PLRDiff \cite{rui2024unsupervised}   & \textbf{32.69}  & 0.8983  & 6.0539  & 1.8370  & \underline{0.9220}  \\          
      & PSDip \cite{rui2024variational}  & 30.71  & 0.8867  & 6.4948  & 1.9637  & 0.9069  \\
      & TTST \cite{xiao2024ttst}  & 32.48  & \textbf{0.9103}  & \underline{5.1607}  & \underline{1.8224}  & \textbf{0.9264}  \\
      & \textbf{Ours} & \textbf{32.69}  & \underline{0.9102}  & \textbf{5.0876}  & \textbf{1.7676}  & 0.9175  \\
\hline
\end{tabular}
}
\label{tab:4_20}
\end{table}

\begin{table}[ht]
\caption{Quantitative results for HSI pansharpening ($\times8$). \textbf{Best} and \underline{second-best} values are highlighted.}
\centering
\resizebox{\columnwidth}{!}{
\begin{tabular}{c|c|c|c|c|c|c}
\hline
\textbf{Dataset} & \textbf{Method}  & \textbf{PSNR} $\uparrow$ & \textbf{SSIM} $\uparrow$ & \textbf{SAM} $\downarrow$ & \textbf{ERGAS} $\downarrow$ & \textbf{SCC} $\uparrow$ \\ 
\hline
      & DBDENet \cite{qu2021dual}   & 22.84  & 0.6945  & 8.5207  & 11.3979  & 0.7971  \\
      & DDLPS  \cite{li2019ddlps}  & 22.27  & 0.5740  & 6.9539  & 17.5198  & 0.7401  \\ 
      & DHP-DARN \cite{zheng2020hyperspectral}  & 28.85  & 0.8204  & 4.9084  & 2.8164  & 0.8727  \\
      & DIP-HyperKite \cite{gedara2021hyperspectral}  & 30.24  & 0.8586  & 4.8305  & 2.1305  & 0.8895  \\
      & DMLD-Net \cite{zhang2023learning}   & 26.87  & 0.7971  & 6.5379  & 3.7552  & 0.8771  \\
\textbf{Botswana} & GPPNN \cite{xu2021deep} & 26.44  & 0.8250  & 8.6439  & 3.8965  & 0.8874  \\
      & GS \cite{laben2000process}  & 10.19  & 0.4745  & 15.5523  & 66.1761  & 0.7697  \\
      & GSA \cite{aiazzi2007improving}  & 23.80  & 0.6297  & 6.2035  & 11.6626  & 0.7919  \\
      & Indusion \cite{khan2008indusion}  & 15.30  & 0.6972  & 5.4225  & 9.7633  & 0.8573  \\
      & PLRDiff \cite{rui2024unsupervised}   & 17.84  & 0.2932  & 15.1475  & 9.0164  & 0.3124  \\            
      & PSDip \cite{rui2024variational}  & 23.67  & 0.6748  & 7.0314  & 3.9725  & 0.7281  \\
      & TTST \cite{xiao2024ttst}  & \underline{30.73}  & \underline{0.8701}  & \underline{4.5100}  & \underline{2.1297}  & \underline{0.9040}  \\
      & \textbf{Ours}   & \textbf{30.92}  & \textbf{0.8918}  & \textbf{4.0053}  & \textbf{1.9397}  & \textbf{0.9147}  \\
\hline
      & DBDENet \cite{qu2021dual}   & 23.63  & 0.5840  & 18.5981  & 6.7944  & 0.5842  \\
      & DDLPS  \cite{li2019ddlps}  & 27.52  & 0.7562  & \textbf{10.1478}  & 4.3781  & 0.7595  \\             
      & DHP-DARN \cite{zheng2020hyperspectral}  & 26.70  & 0.7442  & 12.4018  & 4.7917  & 0.7328  \\
      & DIP-HyperKite \cite{gedara2021hyperspectral}   & 26.48  & 0.7321  & 12.5846  & 4.9198  & 0.7060  \\
      & DMLD-Net \cite{zhang2023learning}  & 26.03  & 0.7652  & 16.8061  & 5.1832  & 0.7178  \\
\textbf{PaviaC} & GPPNN \cite{xu2021deep}  & 27.37  & 0.7763  & 11.2643  & 4.4916  & 0.7813  \\
      & GS \cite{laben2000process} & 24.19  & 0.4665  & 18.7760  & 6.7900  & 0.7755  \\
      & GSA \cite{aiazzi2007improving} & 25.30  & 0.6690  & 10.4678  & 5.6518  & 0.7452  \\
      & Indusion \cite{khan2008indusion} & 25.84  & 0.7166  & \underline{10.4645}  & 5.8683  & 0.7396  \\
      & PLRDiff \cite{rui2024unsupervised}  & 27.39  & 0.7489  & 11.6904  & 4.6245  & 0.7498  \\    
      & PSDip \cite{rui2024variational}   & 21.98  & 0.5495  & 18.9236  & 8.3069  & 0.5144  \\
      & TTST \cite{xiao2024ttst}  & \underline{28.14}  & \underline{0.8063}  & 10.8695  & \underline{4.1102}  & \underline{0.7944}  \\
      & \textbf{Ours}  & \textbf{29.22}  & \textbf{0.8499}  & 11.7926  & \textbf{3.6820}  & \textbf{0.8178}  \\
\hline
      & DBDENet \cite{qu2021dual} & 28.84  & 0.8712  & 6.5032  & 2.7593  & 0.8693  \\
      & DDLPS  \cite{li2019ddlps}   & 27.81  & 0.7894  & 7.1405  & 2.9323  & 0.7913  \\    
      & DHP-DARN \cite{zheng2020hyperspectral} & 29.27  & 0.8284  & 6.8826  & 2.5350  & 0.8231  \\
      & DIP-HyperKite \cite{gedara2021hyperspectral}   & 29.30  & 0.8385  & 6.0972  & 2.5114  & 0.8296  \\
      & DMLD-Net \cite{zhang2023learning} & 28.66  & 0.8584  & 6.8624  & 2.7985  & 0.8572  \\
\textbf{PaviaU} & GPPNN \cite{xu2021deep} & 29.86  & 0.8811  & 6.0788  & 2.4812  & 0.8775  \\
      & GS \cite{laben2000process}  & 24.56  & 0.6269  & 12.1452  & 4.4184  & 0.7984  \\
      & GSA \cite{aiazzi2007improving}  & 26.47  & 0.7514  & 7.2522  & 3.4839  & 0.7775  \\
      & Indusion \cite{khan2008indusion} & 25.82  & 0.7513  & 7.8229  & 4.1539  & 0.7612  \\
      & PLRDiff \cite{rui2024unsupervised}   & 28.57  & 0.7925  & 7.5217  & 2.9453  & 0.7751  \\      
      & PSDip \cite{rui2024variational}  & 21.37  & 0.6068  & 12.2495  & 6.6578  & 0.5164  \\
      & TTST \cite{xiao2024ttst}  & \underline{31.00}  & \underline{0.8931}  & \underline{5.1492}  & \underline{2.1190}  & \textbf{0.8979}  \\
      & \textbf{Ours}  & \textbf{31.61}  & \textbf{0.8982}  & \textbf{5.1381}  & \textbf{2.0157}  & \underline{0.8973}  \\
\hline
\end{tabular}
}
\label{tab:8_4}
\end{table}

\section{Experiment}

This section presents the experimental evaluation of the proposed THAT method on both airborne and earth observation satellite datasets. All datasets are normalized to the range \([0,1]\), and the central region of each dataset is cropped to obtain an HR-HSI of size \(256 \times 256\). The LR-HSI and HR-PCI are generated following Wald’s protocol~\cite{wald1997fusion, ranchin2000fusion}, a widely adopted standard in image fusion tasks. For LR-HSI generation, the HR-HSI is first spatially blurred using a \(20 \times 20\) Gaussian filter and then downsampled by a factor of 2 or 4 to simulate low-resolution hyperspectral data. Alternatively, in some cases, a \(4 \times 4\) Gaussian filter is applied, followed by downsampling by a factor of 8, to further reduce spatial resolution. The HR-PCI is obtained by averaging the visible bands of the HR-HSI, providing a high-resolution panchromatic counterpart for the fusion process. 

\subsection{Datasets}  
We evaluate hyperspectral pansharpening on three publicly available datasets. The Pavia Centre (PaviaC) and Pavia University (PaviaU) datasets \cite{plaza2009recent, fauvel2008spectral} were captured by the ROSIS sensor over Pavia, Italy, and contain 102 spectral bands (430–860 nm) with a spatial resolution of 1.3 m. PaviaC covers an area of $1096 \times 715$ pixels, suitable for urban mapping, while PaviaU spans $610 \times 340$ pixels and includes nine land-cover classes. The Botswana dataset \cite{ungar2003overview}, acquired by NASA’s EO-1 Hyperion sensor over the Okavango Delta, has a spatial resolution of 30 m and an image size of $1476 \times 256$. It originally contained 242 bands (400–2500 nm), but was preprocessed to retain 145 cleaned bands for analysis.

\begin{figure*}[htbp]
    \centering
    \includegraphics[width=\textwidth, height=9.5cm]{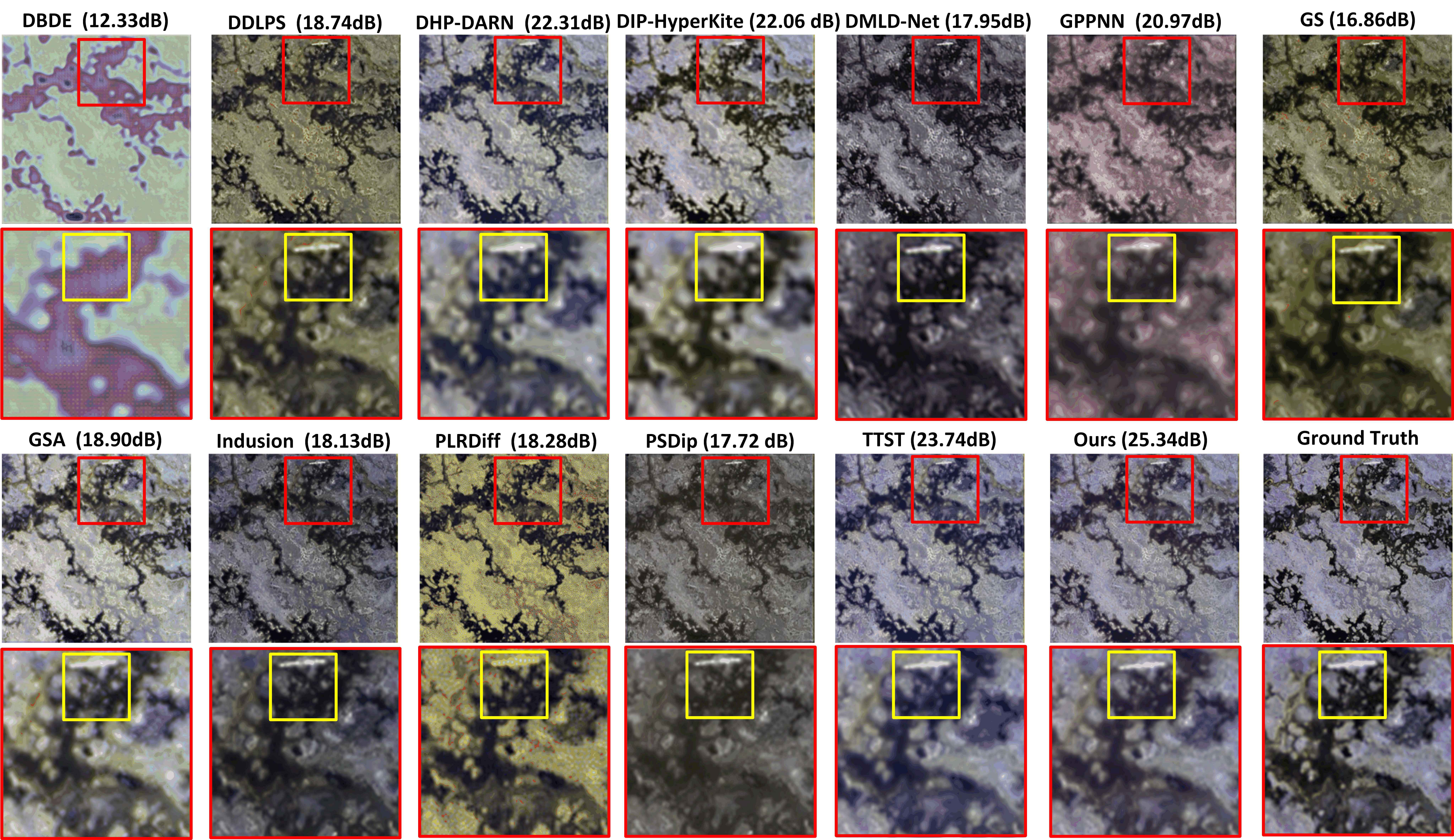}
\caption{Visual results on the Botswana dataset for HSI pansharpening with a $\times2$ scaling factor.} 

    \label{fig:result}
\end{figure*}

\subsection{Evaluation Metrics}
The performance of the proposed THAT method is rigorously evaluated on both airborne and Earth observation satellite datasets, demonstrating its effectiveness in hyperspectral pansharpening across various scenarios. Comparisons are conducted against eleven state-of-the-art methods, including DBDENet \cite{qu2021dual}, DDLPS \cite{li2019ddlps}, DHP-DARN \cite{zheng2020hyperspectral}, DIP-HyperKite \cite{gedara2021hyperspectral}, DMLD-Net \cite{zhang2023learning}, GPPNN \cite{xu2021deep}, GS \cite{laben2000process}, GSA \cite{aiazzi2007improving}, Indusion \cite{khan2008indusion}, PLRDiff \cite{rui2024unsupervised} and PSDip \cite{rui2024variational}.

The performance of the reconstructed HSIs is assessed using five quantitative metrics: Peak Signal-to-Noise Ratio (PSNR), Structural Similarity Index Measure (SSIM), Spectral Angle Mapper (SAM), Error Relative Global Dimension Synthesis (ERGAS) and Spatial Correlation Coefficient (SCC).

\subsection{Implementation Details}

THAT is implemented by the PyTorch framework and trained in an iterative alternating manner on a single NVIDIA GeForce RTX 3090 24-GB graphics processor. The learning rate was initialized to \( 5 \times 10^{-4} \) and reduced by half after every 20 epochs, following a step decay strategy. The models were trained for 50 epochs using the Adam optimizer with a weight decay of 0. The channel number in THAT is set to 180. In WSA, the number of multi-head self-attention is 6. The batch size was set to 2, and the L1 loss function was employed for supervision. 

\begin{figure}[t]
    \centering
    \resizebox{\columnwidth}{!}{
        \includegraphics{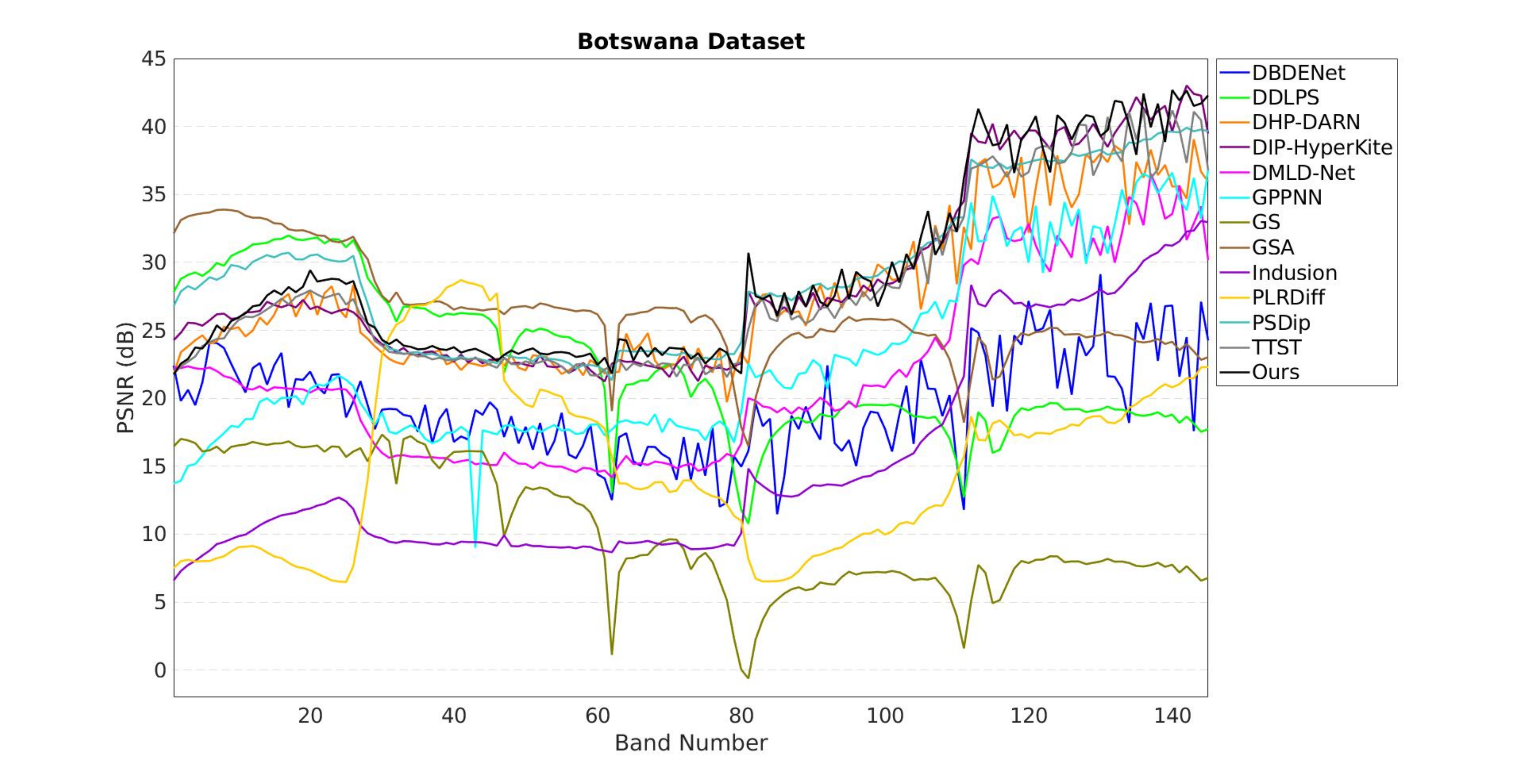}
    }
\caption{Band-wise PSNR comparison for HSI pansharpening ($\times2$) on the Botswana dataset.}

    \label{fig:bo_2}
\end{figure}

\begin{figure}[t]
    \centering
    \resizebox{\columnwidth}{!}{
        \includegraphics{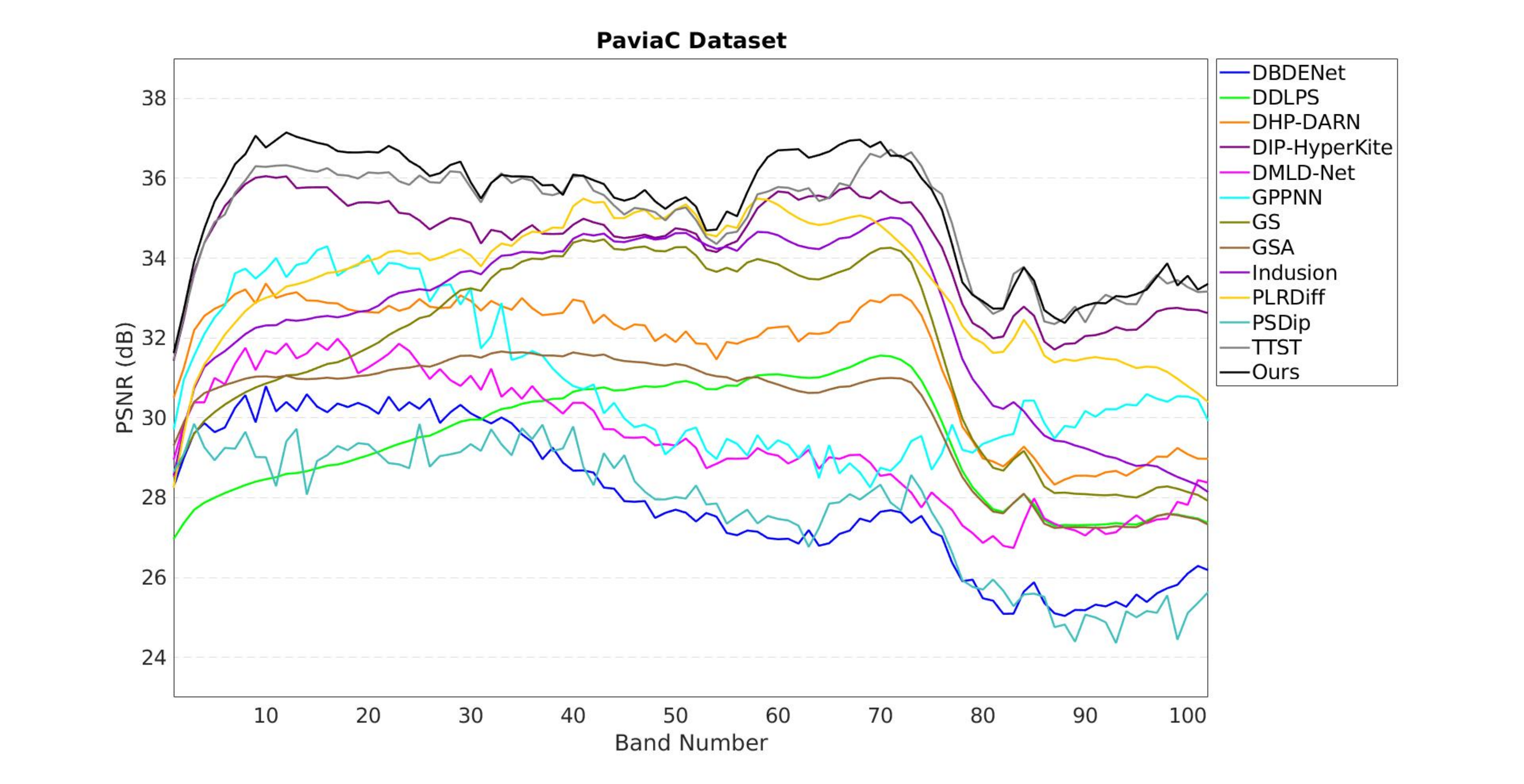}
    }
\caption{Band-wise PSNR comparison for HSI pansharpening ($\times2$) on the PaviaC dataset.}

    \label{fig:pc_2}
\end{figure}

\section{Results Discussion}
\subsection{Results for Hyperspectral Pansharpening}  
\begin{figure}[t]
    \centering
    \resizebox{\columnwidth}{!}{
        \includegraphics{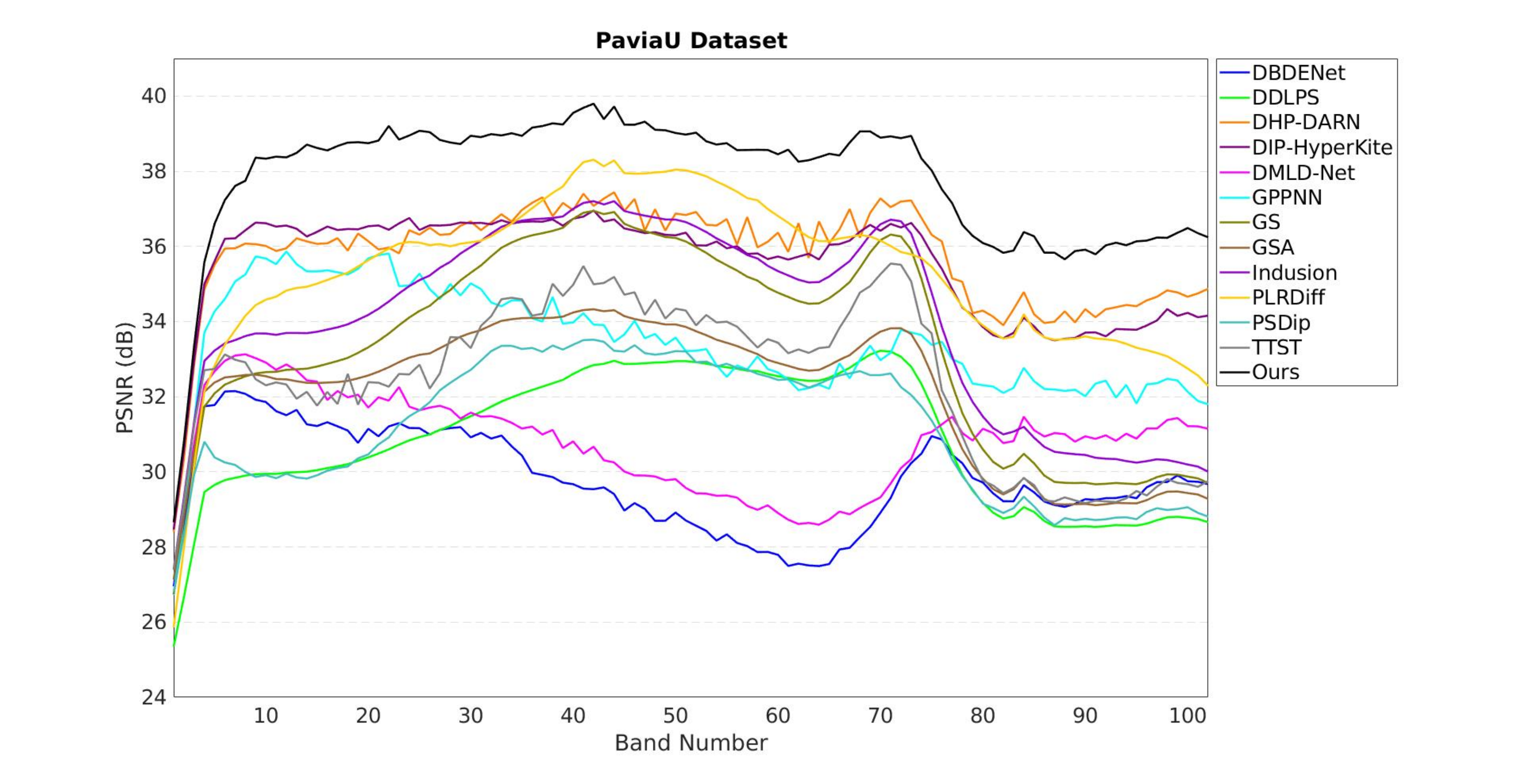}
    }
\caption{Band-wise PSNR comparison for HSI pansharpening ($\times2$) on the PaviaU dataset.}

    \label{fig:pu_2}
\end{figure}

Our proposed method demonstrates state-of-the-art performance across multiple upscaling factors ($\times2$, $\times4$, and $\times8$), consistently outperforming both traditional and deep learning-based approaches (Table~\ref{tab:2_20}, Table~\ref{tab:4_20}, and Table~\ref{tab:8_4}). Fig.~\ref{fig:bo_2}, Fig.~\ref{fig:pc_2}, and Fig.~\ref{fig:pu_2} visualize the PSNR distribution across spectral bands for $\times2$ pansharpening on the Botswana, PaviaU, and PaviaC datasets. Our method consistently achieves the highest PSNR across most spectral bands, demonstrating superior spectral fidelity and noise robustness. Further qualitative results are illustrated in Fig.~\ref{fig:result}, which displays $\times2$ pansharpening outputs on the Botswana dataset using three selected spectral bands. Our method effectively maintains spatial structures and spectral coherence, while reducing spectral distortions. The higher PSNR values further validate its ability to generate high-quality hyperspectral pansharpened images, highlighting its reliability.

\begin{table}[t!]
\centering
\caption{Quantitative evaluation of HR-PCI, PTSA and MVFN on the HSI pansharpening with $\times2$, $\times4$ and $\times4$ scaling factor (SF). }
\label{tab:ablation}
\resizebox{\columnwidth}{!}{
\begin{tabular}{@{}cclllllll@{}}
\toprule
\multirow{2}{*}{\textbf{Module}} & \multirow{2}{*}{\textbf{SF}} & \multicolumn{1}{c}{\multirow{2}{*}{\textbf{Dataset}}} & \multicolumn{3}{c}{\textbf{Module (\ding{51})}}                                                                          & \multicolumn{3}{c}{\textbf{Module (\ding{55})}}                                                                          \\ \cmidrule(l){4-9} 
                                 &                                          & \multicolumn{1}{c}{}                                  & \multicolumn{1}{c}{\textbf{PSNR ↑}} & \multicolumn{1}{c}{\textbf{SSIM ↑}} & \multicolumn{1}{c}{\textbf{ERGAS ↓}} & \multicolumn{1}{c}{\textbf{PSNR ↑}} & \multicolumn{1}{c}{\textbf{SSIM ↑}} & \multicolumn{1}{c}{\textbf{ERGAS ↓}} \\ \midrule
\multirow{9}{*}{\textbf{HR-PCI}} & \multirow{3}{*}{x2}                      & Botswana                                              & \textbf{29.18}                      & \textbf{0.9084}                     & \textbf{2.0754}                      & 28.45                               & 0.8579                              & 2.3665                               \\
                                 &                                          & PaviaC                                                & \textbf{35.29}                      & \textbf{0.9574}                     & \textbf{1.8838}                      & 31.84                               & 0.9224                              & 2.7675                               \\
                                 &                                          & PaviaU                                                & \textbf{37.82}                      & \textbf{0.9632}                     & \textbf{1.0039}                      & 33.24                               & 0.9266                              & 1.6454                               \\ \cmidrule(l){2-9} 
                                 & \multirow{3}{*}{x4}                      & Botswana                                              & \textbf{29.34}                      & \textbf{0.8728}                     & \textbf{2.3373}                      & 26.52                               & 0.6889                              & 2.9287                               \\
                                 &                                          & PaviaC                                                & \textbf{32.43}                      & \textbf{0.9157}                     & \textbf{2.5841}                      & 26.59                               & 0.7117                              & 5.0114                               \\
                                 &                                          & PaviaU                                                & \textbf{32.69}                      & \textbf{0.9102}                     & \textbf{1.7670}                      & 27.22                               & 0.7446                              & 3.2611                               \\ \cmidrule(l){2-9} 
                                 & \multirow{3}{*}{x8}                      & Botswana                                              & \textbf{30.92}                      & \textbf{0.8918}                     & \textbf{1.9397}                      & 24.77                               & 0.4919                              & 3.7049                               \\
                                 &                                          & PaviaC                                                & \textbf{29.22}                      & \textbf{0.8499}                     & \textbf{3.6820}                      & 23.40                               & 0.4872                              & 7.2017                               \\
                                 &                                          & PaviaU                                                & \textbf{31.61}                      & \textbf{0.8982}                     & \textbf{2.0157}                      & 24.00                               & 0.5603                              & 4.7333                               \\ \midrule
\multirow{9}{*}{\textbf{PTSA}}   & \multirow{3}{*}{x2}                      & Botswana                                              & \textbf{29.18}                      & \textbf{0.9084}                     & \textbf{2.0754}                      & 28.08                               & 0.8846                              & 2.5061                               \\
                                 &                                          & PaviaC                                                & \textbf{35.29}                      & \textbf{0.9574}                     & \textbf{1.8838}                      & 34.94                               & 0.9524                              & 1.9735                               \\
                                 &                                          & PaviaU                                                & \textbf{37.82}                      & \textbf{0.9632}                     & \textbf{1.0039}                      & 37.71                               & 0.9626                              & 1.0328                               \\ \cmidrule(l){2-9} 
                                 & \multirow{3}{*}{x4}                      & Botswana                                              & \textbf{29.34}                      & 0.8728                              & 2.3373                               & 28.89                               & \textbf{0.8738}                     & \textbf{2.2224}                      \\
                                 &                                          & PaviaC                                                & \textbf{32.43}                      & \textbf{0.9157}                     & \textbf{2.5841}                      & 32.19                               & 0.9116                              & 2.6717                               \\
                                 &                                          & PaviaU                                                & \textbf{32.69}                      & 0.9102                              & \textbf{1.7670}                      & 32.45                               & \textbf{0.9119}                     & 1.8616                               \\ \cmidrule(l){2-9} 
                                 & \multirow{3}{*}{x8}                      & Botswana                                              & \textbf{30.92}                      & \textbf{0.8918}                     & \textbf{1.9397}                      & 30.28                      & 0.8623                              & 2.2157                               \\
                                 &                                          & PaviaC                                                & \textbf{29.22}                      & \textbf{0.8499}                     & \textbf{3.6820}                       & 28.34                               & 0.8355                              & 4.0366                               \\
                                 &                                          & PaviaU                                                 & \textbf{31.61}                      & \textbf{0.8982}                     & \textbf{2.0157}                          & 31.41                               & 0.8970                              & 2.0812                               \\ \midrule
\multirow{9}{*}{\textbf{MVFN}}   & \multirow{3}{*}{x2}                      & Botswana                                              & \textbf{29.18}                      & \textbf{0.9084}                     & \textbf{2.0754}                      & 28.24                               & 0.8862                              & 2.6411                               \\
                                 &                                          & PaviaC                                                & \textbf{35.29}                      & \textbf{0.9574}                     & \textbf{1.8838}                      & 34.13                               & 0.9501                              & 2.1607                               \\
                                 &                                          & PaviaU                                                & \textbf{37.82}                      & \textbf{0.9632}                     & \textbf{1.0039}                      & 37.49                               & 0.9609                              & 1.0624                               \\ \cmidrule(l){2-9} 
                                 & \multirow{3}{*}{x4}                      & Botswana                                              & \textbf{29.34}                      & \textbf{0.8728}                     & \textbf{2.3373}                      & 28.68                               & 0.8523                              & 2.3870                               \\
                                 &                                          & PaviaC                                                & \textbf{32.43}                      & \textbf{0.9157}                     & \textbf{2.5841}                      & 32.14                               & 0.9061                              & 2.6633                               \\
                                 &                                          & PaviaU                                                & \textbf{32.69}                      & \textbf{0.9102}                     & \textbf{1.7670}                      & 32.51                               & 0.7446                              & 1.7820                               \\ \cmidrule(l){2-9} 
                                 & \multirow{3}{*}{x8}                      & Botswana                                              & \textbf{30.92}                      & \textbf{0.8918}                     & \textbf{1.9397}                      & 30.46                               & 0.8743                              & 2.0099                               \\
                                 &                                          & PaviaC                                                & \textbf{29.22}                      & \textbf{0.8499}                     & \textbf{3.6820}                      & 28.59                               & 0.8303                              & 3.9142                               \\
                                 &                                          & PaviaU                                                & \textbf{31.61}                      & \textbf{0.8982}                     & \textbf{2.0157}                      & 31.54                               & 0.8947                              & 2.0157                               \\ \bottomrule
\end{tabular}
}
\end{table}

\begin{table}[t]
\small
\centering
\caption{Performance comparison with and without the module across datasets and scale factors (SF). The best results are in bold.}
\resizebox{\columnwidth}{!}{
\begin{tabular}{ccccc ccc}
\toprule
SF & Dataset & \multicolumn{3}{c}{MVFN} & \multicolumn{3}{c}{MFL} \\
\cmidrule(lr){3-5} \cmidrule(lr){6-8}
& & PSNR $\uparrow$ & SSIM $\uparrow$ & ERGAS $\downarrow$ & PSNR $\uparrow$ & SSIM $\uparrow$ & ERGAS $\downarrow$ \\
\midrule
\multirow{3}{*}{$\times$2}
& Botswana & \textbf{29.18} & \textbf{0.9084} & \textbf{2.0754} & 28.18 & 0.8806 & 2.3206 \\
& PaviaC   & \textbf{35.29} & \textbf{0.9574} & \textbf{1.8838} & 34.64 & 0.9518 & 2.0318 \\
& PaviaU   & \textbf{37.82} & \textbf{0.9632} & \textbf{1.0039} & 36.99 & 0.9585 & 1.1018 \\
\midrule
\multirow{3}{*}{$\times$4}
& Botswana & \textbf{29.34} & \textbf{0.8728} & \textbf{2.3373} & 28.82 & 0.8554 & 2.5417 \\
& PaviaC   & \textbf{32.43} & \textbf{0.9157} & \textbf{2.5841} & 31.98 & 0.9103 & 2.6950 \\
& PaviaU   & \textbf{32.69} & \textbf{0.9102} & \textbf{1.7670} & 31.83 & 0.9012 & 1.9686 \\
\midrule
\multirow{3}{*}{$\times$8}
& Botswana & \textbf{30.92} & \textbf{0.8918} & \textbf{1.9397} & 29.91 & 0.8651 & 2.2035 \\
& PaviaC   & \textbf{29.22} & \textbf{0.8499} & \textbf{3.6820} & 28.24 & 0.8227 & 4.0404 \\
& PaviaU   & \textbf{31.61} & \textbf{0.8982} & \textbf{2.0157} & 30.51 & 0.8655 & 2.3654 \\
\bottomrule
\end{tabular}}
\label{tab:module_ablation}
\end{table}

\subsection{Ablation Study}
\subsubsection{Effects of HR-PCI Fusion on HSI Pansharpening}
Our experiments (Table~\ref{tab:ablation}) show that HR-PCI fusion consistently improves hyperspectral pansharpening across all scales (\(\times2\), \(\times4\), \(\times8\)), boosting PSNR, SSIM, and reducing ERGAS. It achieves up to 7.61 dB PSNR gain (PaviaU) and significant ERGAS reduction (49\% in PaviaC), demonstrating strong spectral–spatial fidelity and robustness under extreme upscaling (\(\times8\)).
\subsubsection{Effectiveness of PTSA}

PTSA dynamically refines self-attention by emphasizing informative tokens and filtering redundancy, leading to consistent gains in PSNR, SSIM, and ERGAS (Table~\ref{tab:ablation}). It yields up to 0.88 dB PSNR improvement (PaviaC, \(\times8\)) and ERGAS reduction (4.55\% in PaviaC, \(\times2\)), confirming its effectiveness in enhancing spectral–spatial consistency across all scales.

\subsubsection{Effectiveness of MVFN}

MVFN captures hierarchical spectral–spatial dependencies to enhance feature representation. As shown in Table~\ref{tab:ablation}, it consistently boosts PSNR, SSIM, and reduces ERGAS across all scales. To demonstrate the importance of high-frequency feature learning, we replace MVFN with the Multi-scale Feed-forward Layer (MFL) \cite{xiao2024ttst}. MVFN is designed to enhance high-frequency detail by modeling spectral–spatial variance across multiple scales. Unlike MFL, which applies parallel depth-wise convolutions followed by simple concatenation, MVFN introduces variance modeling and adaptive pooling for richer and more informative feature representation. As shown in Table~\ref{tab:module_ablation}, MVFN consistently outperforms MFL across all datasets and scale factors.

\subsection{Complexity Analysis}
Table~\ref{tab:para} compares the parameter count and FLOPs of representative methods on the Botswana dataset. While PLRDiff \cite{rui2024unsupervised} is the most computationally intensive, our method achieves a favorable trade-off, with moderate complexity (1.45 M parameters and 78.42 G FLOPs) compared to other efficient models like TTST \cite{xiao2024ttst}.

\begin{table}[t!]
\centering
\caption{The Parameters and FLOPs of different deep learning models on the Botswana Dataset.}

\begin{tabular}{c|c|c}
\hline
Method         & Parameters & FLOPs \\ \hline
DBDENet\cite{qu2021dual}        & 1.32 M    & 117.76 G       \\
DHP-DARN \cite{zheng2020hyperspectral}      & 0.47 M     & 30.38 G       \\
DIP-HyperKite \cite{gedara2021hyperspectral}  & 0.27 M     & 426.90 G       \\
DMLD-Net \cite{zhang2023learning}        & 0.49 M     & 18.55 G      \\
GPPNN \cite{xu2021deep}        & 4.31 M     & 196.05 G       \\
HyperPNN \cite{he2019hyperpnn}       & 0.14 M     & 9.07 G       \\
PLRDiff \cite{rui2024unsupervised}       & 391.05 M   & 22.43 T       \\
TTST \cite{xiao2024ttst}         & 1.32 M    & 92.17 G       \\ 
\textbf{Ours}         & 1.45 M    & 78.42 G       \\ \hline
\end{tabular}
\label{tab:para}
\end{table}

\section{Conclusion}

We proposed the Token-wise High-frequency Augmentation Transformer (THAT) for hyperspectral pansharpening, targeting key limitations in token redundancy and inadequate multi-scale feature modeling. THAT comprises three core components: (1) Pivotal Token Selective Attention (PTSA), which prioritizes informative tokens and suppresses redundancy to enhance self-attention and (2) a Multi-level Variance-aware Feed-forward Network (MVFN), which strengthens high-frequency detail learning through hierarchical variance-guided representation. Extensive experiments on benchmarks show THAT achieves superior results.

\bibliographystyle{IEEEtran}
\bibliography{main}

\end{document}